\documentclass[conference]{IEEEtran}
\IEEEoverridecommandlockouts
\usepackage{cite}
\usepackage{amsmath,amssymb,amsfonts}
\usepackage{algorithmic}
\usepackage{graphicx}
\usepackage{textcomp}
\usepackage{xcolor}
\usepackage{adjustbox}
\def\BibTeX{{\rm B\kern-.05em{\sc i\kern-.025em b}\kern-.08em
    T\kern-.1667em\lower.7ex\hbox{E}\kern-.125emX}}
\begin{document}

\title{Using Texture to Classify \\ Forests Separately from Vegetation\\
}

\makeatletter
\newcommand{\linebreakand}{%
  \end{@IEEEauthorhalign}
  \hfill\mbox{}\par
  \mbox{}\hfill\begin{@IEEEauthorhalign}
}
\makeatother

\author{\IEEEauthorblockN{1\textsuperscript{st} David R. Treadwell IV}
\IEEEauthorblockA{\textit{Computer Science} \\
\textit{Northeastern University}\\
Seattle, WA, USA \\
treadwell.d@northeastern.edu}
\and
\IEEEauthorblockN{2\textsuperscript{nd} Derek Jacoby}
\IEEEauthorblockA{\textit{Computer Science} \\
\textit{University of Victoria}\\
Victoria, BC, Canada \\
https://orcid.org/0000-0002-1552-7484}
\and
\IEEEauthorblockN{3\textsuperscript{rd} Will Parkinson}
\IEEEauthorblockA{\textit{Earth Daily Analytics, Inc} \\
Vancouver, BC, Canada \\
Will.Parkinson@earthdaily.com}
\linebreakand
\IEEEauthorblockN{4\textsuperscript{th} Bruce Maxwell}
\IEEEauthorblockA{\textit{Computer Science} \\
\textit{Northeastern University}\\
Seattle, Wa, USA \\
b.maxwell@northeastern.edu}
\and
\IEEEauthorblockN{5\textsuperscript{th} Yvonne Coady}
\IEEEauthorblockA{\textit{Computer Science} \\
\textit{University of Victoria}\\
Victoria, BC, Canada \\
ycoady@uvic.ca}

}

\maketitle

\begin{abstract}
Identifying terrain within satellite image data is a key issue in geographical information sciences, with numerous environmental and safety implications. Many techniques exist to derive classifications from spectral data captured by satellites. However, the ability to reliably classify vegetation remains a challenge. In particular, no precise methods exist for classifying forest vs. non-forest vegetation in high-level satellite images. This paper provides an initial proposal for a static, algorithmic process to identify forest regions in satellite image data through texture features created from detected edges and the NDVI ratio captured by Sentinel-2 satellite images. With strong initial results, this paper also identifies the next steps to improve the accuracy of the classification and verification processes. \\
\end{abstract}

\begin{IEEEkeywords}
texture, image processing, satellite images, forest, vegetation, terrain classification
\end{IEEEkeywords}

\section{Introduction}

The ability to classify regions of terrain from satellite images is a critical step in monitoring the health of ecological systems. Algorithms exist to help parse the visual bands captured by satellites, such as the Level-2A Algorithm to classify terrain properties in Sentinel-2 image data. \cite{noauthor_level-2a_nodate} However, a missing component of these image processing algorithms is the ability to parse vegetation types - specifically, the ability to separate forest areas from non-forest vegetation. Such classifications can improve land usage monitoring and aid in identifying ecological risks, which in turn can help prevent natural disasters and increase emergency preparedness.

Currently, techniques exist to identify areas of trees at the pixel level, but these are primarily designed for close-to-ground aerial images and urban environments. This paper seeks to demonstrate that by utilizing both textural and spectral data from Sentinel-2 satellite images, the binary classification of forest vs. non-forest vegetation can be performed. The static classification algorithm uses a texture mask derived from edges within an RGB image, combined with the NDVI image, to classify forest regions with high accuracy when evaluated against alternative methods. Compared to more complex classification techniques, this paper presents evidence that a simple, easy-to-understand static algorithm that combines textural and spectral data can produce high-quality results. 

\section{Related Work}

\subsection{Satellite imagery and preparation}

The European Space Agency (ESA) has been making Sentinel-2 data available since 2014 and offers a platform for freely downloading corrected data. \cite{gascon_copernicus_2017} In this study the imagery is instead downloaded from the Earth Daily Analytics platform on Amazon Web Services. For the moment, the platform offers imagery corrected through the pipeline offered by the ESA, but there are many other emerging options. One of the authors participated in a study of correction techniques for certain classes of coastal waters \cite{giannini_performance_2021} and it is clear that in specific cases alternative correction mechanisms are preferred. The primary element of correction is taking the top-of-atmosphere signal from the satellite and mapping it to a bottom-of-atmosphere estimate. Essentially, the goal is to cleanly remove the effects of the atmosphere from the data. There are both fixed and adaptive methods to perform this correction. The default atmospheric correction mechanism performed by ESA, called sen2cor, is a good general correction. \cite{main-knorn_sen2cor_2017} Other methods, many involving neural networks, provide superior results in specific situations. \cite{li_assessment_2023} In future work, the investigation of correction methods for forest identification will be further investigated.

\subsection{Texture Detection}

Texture, alongside spectral and contextual information, is a well-established feature in segmenting and classifying images. \cite{haralick_textural_1973} Many techniques for utilizing texture-based features as the primary algorithmic classifier exist, as there are many ways to identify spatial and statistical patterns within an image. The conventional approach uses texture independent of spectral and contextual information by evaluating it in grey-tone images. \cite{haralick_textural_1973} This can be seen when using a feature bank of textures based on image statistics \cite{haralick_textural_1973} or other techniques such as a discrete wavelet transform that decomposes an image into sub-bands that are used for feature extraction and classification. \cite{arivazhagan_texture_2003} Texture can also be a useful feature for machine learning classification, such as using local texture descriptors alongside a KNN classifier, which has shown promise in terrain segmentation and classification. \cite{suruliandi_texture-based_2015}

Texture has been a popular topic of research in terrain classification. Sali and Wolfson showed that texture, via second-order statistics about local pixel neighborhoods, could result in near-perfect segmentation of satellite images. \cite{sali_texture_1992} When classifying areas of terrain, especially related to forest and "natural" vegetation such as meadows and shrubs, previous works have identified the benefits of including texture as a feature in classification \cite{zhang_exploring_2023} or solely relying on texture to train a model. \cite{suruliandi_texture-based_2015} These results are especially positive when compared to areas of planted vegetation such as crops. \cite{zhang_exploring_2023} \cite{suruliandi_texture-based_2015} It has also been shown that combining spectral information with texture features can further enhance the accuracy of classification, \cite{haralick_textural_1973} and that spectral ratios specifically designed to analyze vegetation can provide further improved accuracy – again, especially for forest, meadow, and shrub-like vegetation. \cite{zhang_exploring_2023}

Several methods exist to classify tree vs. non-tree areas at the pixel level. Yang et al. utilized a color, texture, and entropy feature vector to create a pixel-level classifier for trees from aerial imagery. \cite{yang_tree_2009} They achieved 91.7\% accuracy using their classifier in urban areas with a high resolution between 0.5-1 meters per pixel. \cite{yang_tree_2009} However, the authors note that the classifier can struggle with areas like grass and when classifying tree types that the classifier was not trained on. \cite{yang_tree_2009} Similarly, the work of Jain et al. uses a Mask Region-based CNN to classify trees from aerial images taken at a high resolution, with an average precision score of 0.36 and an average recall score of 0.42. \cite{jain_efficient_2019} Bosch builds on the work of Yang et al. through "DetecTree", a binary AdaBoost classifier using a GIST feature vector. \cite{bosch_detectree_2020} DetecTree was able to classify pixels with between 85.98\% and 91.75\% accuracy in testing, depending on the scene. \cite{bosch_detectree_2020} While strong methods exist for classifying trees from aerial images, all of these methods focus on high-resolution scenes near to the earth where there is more detail and individual trees can be detected. This is different than the low-resolution and far-distance satellite image scenes where individual trees cannot be accurately discerned (e.g. forests), which are the goal of this paper.

\section{Methods}
\subsection*{Algorithmic Classification}
To classify regions as forest or non-forest from Sentinel-2 satellite image data, the following process is used as part of a static algorithm:

\subsubsection*{1. Obtain the RGB image from the Sentinel-2 satellite images}
To create the RGB image, the corresponding color bands from the Sentinel-2 images were used: the B2 band for blue, B3 band for green, and B4 band for red. \cite{gisgeography_sentinel_2019} After creating the RGB image, it is transformed from the float range [0.0, 1.0] to a range of [0, 255] integers (8-bit unsigned) by multiplying the float values by 255, then integer-dividing by 1 to round decimal values down to integer values. The images were captured from RGB Sentinel-2 tiles, with 500x500 cropped target pixel regions. The satellite images are captured from a mean altitude of 786km, \cite{noauthor_orbit_nodate} with an orbital swath of 290km. \cite{noauthor_sentinel-2_nodate}

\subsubsection*{2. Take the Laplacian of the image to detect edges}
First, the RGB satellite image is converted to greyscale. Then, a Laplacian edge detection filter is applied, with a kernel size of 5. This generates an image where edges are visible, but areas within the edges are not. The filter is applied using the OpenCV Laplacian function. This function uses a Laplace operator that calculates the second derivatives of pixel neighborhoods about a target pixel. \cite{noauthor_opencv_nodate}

\subsubsection*{3. Close regions of detected edges to create areas of high and low texture, rather than just showing where edges exist}
Using a 5x5 kernel, the regions of the Laplacian image are closed using the OpenCV morphologyEx function. The morphological closing works by sliding the kernel across the image, and first dilating (a target pixel is given a value of 1 if any pixel under the kernel has a value of 1), then eroding (a target pixel is given a value of 1 if all pixels under the kernel have a value of 1). This generates an image where high-texture regions are "activated," while regions of low texture are not. \cite{noauthor_opencv_nodate} This output image serves as the texture mask in classification.

\subsubsection*{4. Apply a vegetation mask on the image to identify candidate areas for forests}
The NDVI (Normalized Difference Vegetation Index) image is calculated in the same way that the Level-2A algorithm calculates it: (B8 - B4)/(B8 + B4). \cite{noauthor_level-2a_nodate} This image shows areas of healthy vegetation that absorb the visible red wavelength but reflect the infrared wavelength (higher values in the ratio), as opposed to regions lacking healthy vegetation where more visible red and less infrared wavelengths are reflected (low values). \cite{noauthor_level-2a_nodate} This image is used as the vegetation mask in classification.

\subsubsection*{5. Combine the texture mask and the vegetation mask to identify areas that are likely forest vegetation}
Each pixel of the output image is iterated over. For the examples within this report, this is the original RGB image, but it could also be the fully classified image from the scl (full terrain classification) data generated by the Sentinel-2 tiles. Pixels that are activated in both the texture mask (value greater than 64) and the NDVI image (value greater than 0.5) are classified as forest-pixels in the output image as forest pixels.

\subsection*{Ground-Truth Image Preparation}
To compute the relative accuracy of the static algorithm, as well as the DetecTree classifications for comparison, the Forest Resources Inventory leaf-on LiDAR Landcover data provided by the Ontario Natural Resources and Forestry Ministry was used. \cite{ontario_forest_nodate2} This data was last updated in April of 2023, compared to the satellite image data from June of 2023, which makes them reasonable to compare. \cite{ontario_forest_nodate2} However, processing was required to try to match the original Ontario data (referred to as "ground-truth" data from here on) with the Sentinel-2 satellite image data.

The sentinel-2 imagery for the test region was provided in a projection appropriate to Northern Ontario (EPSG:32615). Generally, to reduce inaccuracies it is desirable to use a projection that is as specific to the region of interest as possible. However, the more general ground truth data (the Ontario Forest Resources Inventory, FRI) is collected from airborne LiDaR sources and provided in an EPSG:3857 projection. To align the satellite and ground truth imagery, it was necessary to use QGIS to re-project the FRI data into EPSG:32615, and then to extract the portion of the imagery that corresponds to the area of interest. This was FRI data exported and aligned with the Sentinel-2 tile region.

With a ground-truth region similar to the Sentinel-2 tile region, additional steps were taken to obtain the analysis region to compare with. Due to the challenges with aligning ground-truth data and Sentinel-2, two methods were used to obtain an estimated ground-truth scene:

\begin{enumerate}
    \item \textbf{Different Projections} Using an original image of size (1811, 1465, 3) [rows, columns, color channels], the full-size ground-truth map is cropped to resemble the Sentinel-2 image. This corresponds roughly to trimming pure-black pixels from the top, right, and bottom edges of the ground-truth map, though the actual cropping may differ due to slight misalignments resulting from the different projection types. Pure-black ([0, 0, 0] RGB value) pixels are then concatenated horizontally to the left edge of the ground-truth image, to make it a square with the dimensions of (1783, 1783, 3). After cropping the full map to the target scene, the final ground-truth image size in this method is (81, 81, 3).

\begin{figure}[!htbp]
    \centering
    \includegraphics[width=0.75\linewidth]{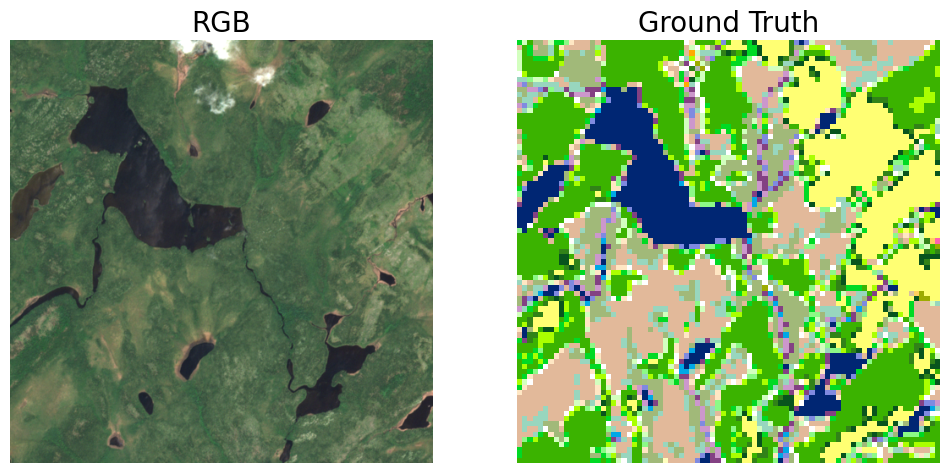}
    \caption{Comparison of RGB image from Sentinel-2 satellite with ground-truth map generated via method 1.}
    \label{fig:gt1}
\end{figure}
    
    \item \textbf{Similar Projections} Using an original image of size (1098, 872, 3), the full-size ground-truth map aligned with the Sentinel-2 image data is used. Similar to method 1, pure-black pixels were concatenated to the left edge to pad the image dimensions to a square of size (1098, 1098, 3). After cropping to the target scene, the final image size in this method is (50, 50, 3).

\begin{figure}[!htbp]
    \centering
    \includegraphics[width=0.75\linewidth]{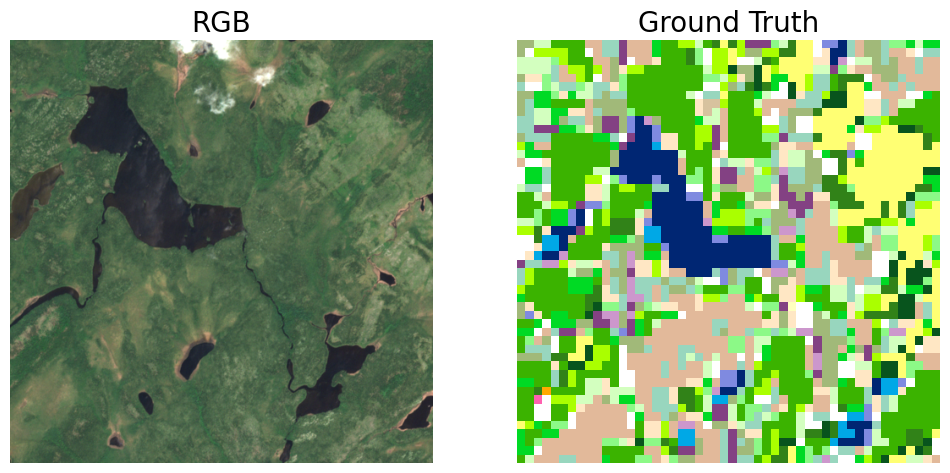}
    \caption{Comparison of RGB image from Sentinel-2 satellite with ground-truth map generated via method 2.}
    \label{fig:gt2}
\end{figure}
    
\end{enumerate}
After preparing the square images, both methods use a ratio of ground-truth size (respective to the method) divided by Sentinel-2 image size. These use the number of rows in the image but could have used the number of columns because the image is a square. This ratio of pixels in the ground-truth image is used to obtain a square scene that roughly matches the target scene from the satellite image. When performing the comparison, for accuracy, these images are also resized to (500, 500, 3) to match the satellite image size, but this resizing is not shown in figures \ref{fig:gt1} or \ref{fig:gt2}. 

To compare the static algorithm's classification and ground-truth image, a pixel-by-pixel comparison of images was performed. Pixels were ignored entirely if the ground-truth image did not classify them, which could manifest as a ground-truth pixel classification of "Other", "Cloud/Shadow", or "Disturbance". Forest pixels were identified as belonging to any of the following classifications: "Sparse Treed", "Treed Upland", "Deciduous Treed", "Mixed Treed", "Coniferous Treed", "Plantations - Treed Cultivated", or "Tallgrass Woodland". The remaining terrain classifications were considered non-forest.

\section{Results}

An example scene with a variety of natural terrain types such as forest, non-forest vegetation, and lake water was used in experiments to test how the static algorithm handled multiple input types. The target scene is the area around Keating Lake in Ontario, Canada, Southeast of Pickle Lake. The masks/steps that lead to a forest identification using this paper's static algorithm are visualized individually, including the output identification image. An image from an open-source tree detector, DetecTree, is also included as a comparison. \cite{bosch_detectree_2020}

\begin{figure}[!htbp]
    \centering
    \includegraphics[width=1\linewidth]{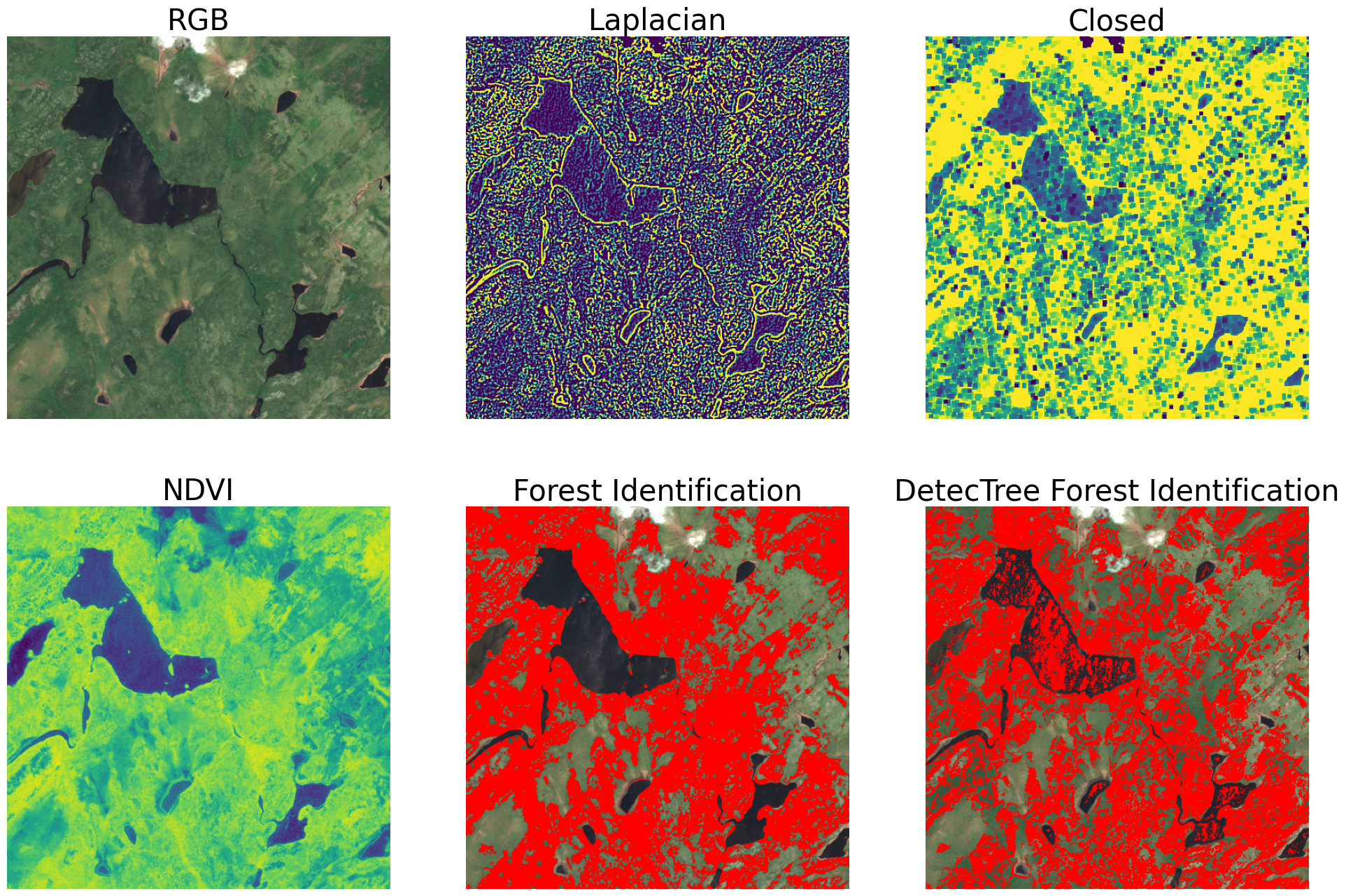}
    \caption{Example of an area with mixed forest and non-forest vegetation, as well as other terrain types such as lake water. The picture shows Keating Lake, Southeast of Pickle Lake in Ontario, Canada.}
    \label{fig:target_scene}
\end{figure}

The images included in the figures are as follows:
\begin{itemize}
    \item \textbf{RGB} The original RGB-generated satellite image, using the appropriate visual bands from Sentinel-2. 
    \item \textbf{Laplacian} A Laplacian filtered image generated by the OpenCV Laplacian function. 
    \item \textbf{Closed} A morphologically closed image generated using the OpenCV morphologyEx function.
    \item \textbf{NDVI} The NDVI image generated using the appropriate visual bands from Sentinel-2.
    \item \textbf{Forest Identification} The output image where forest pixels are highlighted in the original RGB image to demonstrate the classification. Pixels determined to be forest are given a value of [Red: 255, Green: 0, Blue: 0]. Non-forest pixels remain unchanged compared to the original RGB image.
    \item \textbf{DetecTree Forest Identification} The predicted tree regions from the DetecTree classifier. \cite{bosch_detectree_2020} Tree-classified pixels are given a pure-red value as in the Forest Identification image, for comparison between the two methods.
\end{itemize}

The figures show initially promising results, as areas that appear to be forest are highlighted red while non-forest areas are not altered. The combination of texture and spectral features tends to prevent non-forest areas from being identified, while also capturing forest areas. DetecTree provides a more conservative classification but does not always avoid type one or two errors. For instance, parts of the lake are classified as trees, while areas of forest are ignored. DetecTree seems more conservative with classifications near forest edges but also appears to under-classify these forest regions. It is worth mentioning again that the DetecTree model being used in the comparison is designed with aerial images taken closer to the surface in mind, as opposed to satellite images when terrain would be identified with very little closeup detail.

\begin{figure}[!htbp]
    \centering
    \includegraphics[width=1\linewidth]{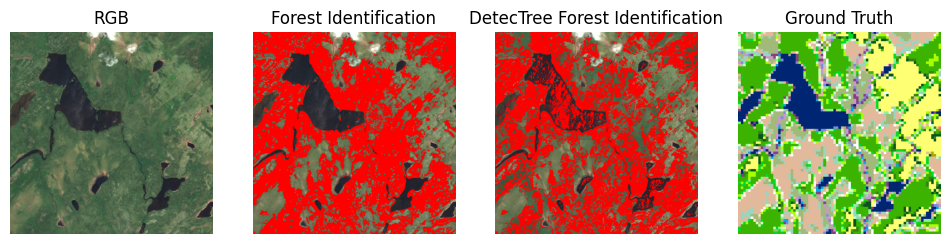}
    \caption{Comparison of the RGB image with the algorithmic classification of pixels as forest, the DetecTree classification of pixels as forest, and the ground-truth image generated through method 1.}
    \label{fig:full_compare_m1}
\end{figure}

\begin{figure}[!htbp]
    \centering
    \includegraphics[width=1\linewidth]{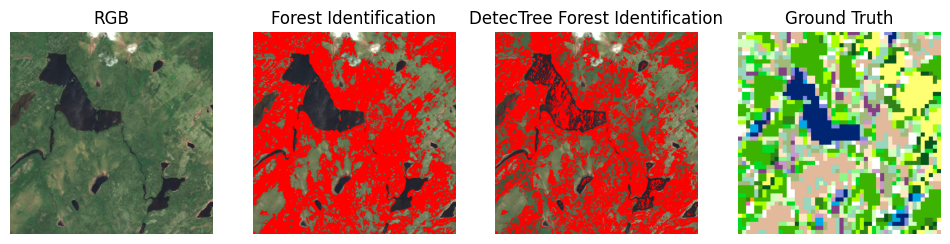}
    \caption{Comparison of the RGB image with the algorithmic classification of pixels as forest, the DetecTree classification of pixels as forest, and the ground-truth image generated through method 2.}
    \label{fig:full_compare_m2}
\end{figure}

While the next component of the paper looks into accuracy values, it is important to keep in mind that, due to the sampling and processing methods, there may be inaccuracies in what is used as ground-truth data. From the comparison of RGB and classification images with the ground-truth image, it becomes clear that there may be differences in pixel precision for the ground-truth image. For example, in figure \ref{fig:full_compare_m1}, the bottom left quadrant seems to switch somewhat randomly between forest and non-forest regions and does not appear to match exactly. 

Viewing the confusion matrix for both methods of ground truth image generation, method 1 (figure \ref{fig:cm_m1}) seems to do a better job of matching the scenes, as there are more true values and less false values between both classification processes than in figure \ref{fig:cm_m2} for method 2. In figure \ref{fig:cm_m1}, the differences between the static algorithm and DetecTree are clearly visible, as the static algorithm predicts fewer false negatives, but also predicts more false positives. This also leads to more true positives, however.

\begin{figure}[!htbp]
    \centering
    \includegraphics[width=1\linewidth]{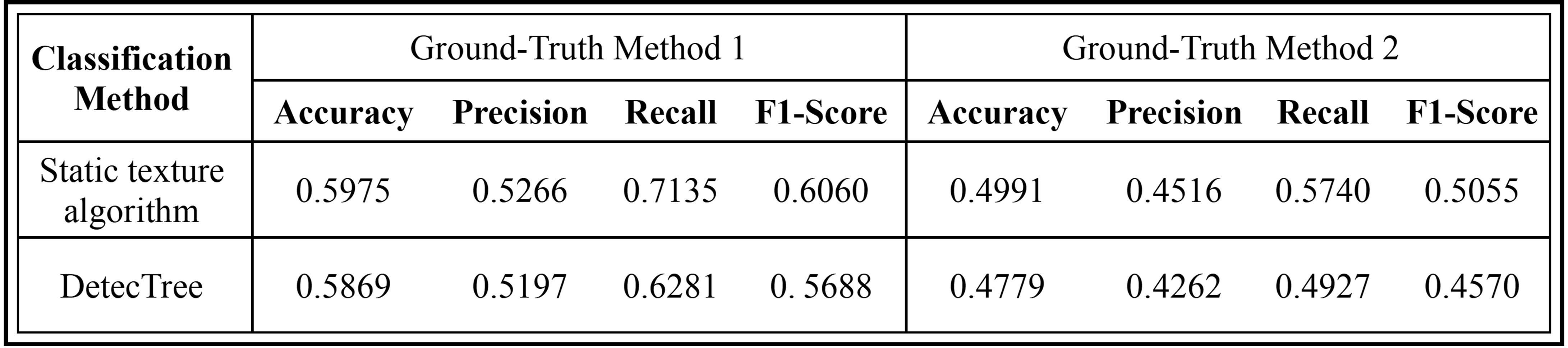}
    \caption{Accuracy, precision, recall, and F1 scores for the static algorithm in this paper and DetecTree, for both ground-truth methods.}
    \label{accuracy_table}
\end{figure}

Based on these tests, figure \ref{accuracy_table} shows that, using ground-truth method 1, the static algorithm accurately classifies 59.75\% of pixels, while DetecTree accurately classifies 58.69\% of pixels. For method 2, the accuracy for both detection processes decreases, with the static algorithm correctly classifying 49.91\% and DetecTree correctly classifying 47.79\% of pixels. From these results, it does appear that the static algorithm is outperforming DetecTree slightly.

\begin{figure}[!htbp]
    \centering
    \includegraphics[width=1\linewidth]{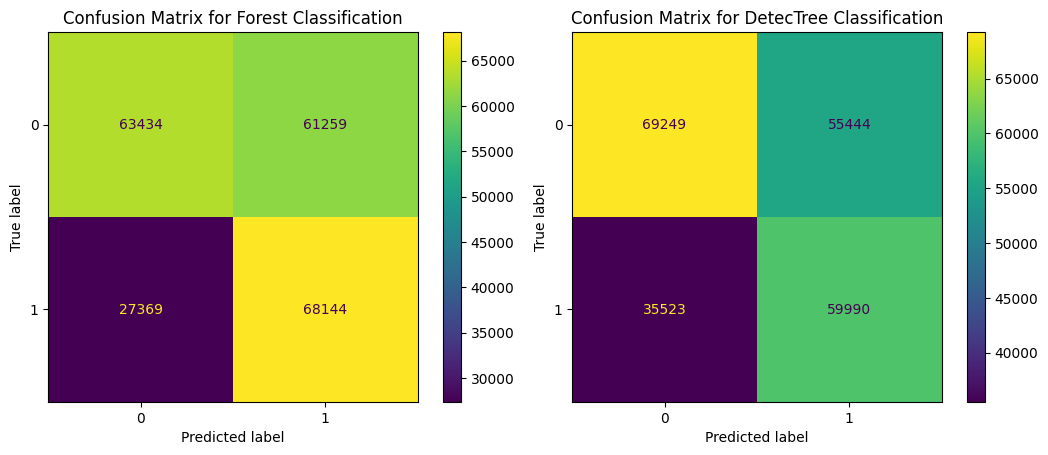}
    \caption{Confusion matrix for both the static algorithm and DetecTree using ground-truth method 1.}
    \label{fig:cm_m1}
\end{figure}

\begin{figure}[!htbp]
    \centering
    \includegraphics[width=1\linewidth]{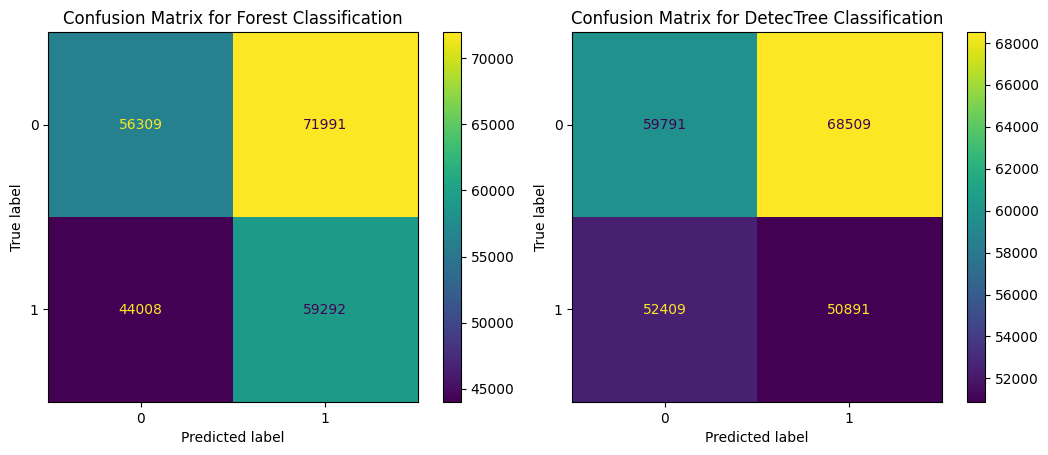}
    \caption{Confusion matrix for both the static algorithm and DetecTree using ground-truth method 2.}
    \label{fig:cm_m2}
\end{figure}

Visually, it is clear that DetecTree is missing areas of forest (this is especially notable in the top-right quadrant), and is not accurately classifying the lake region, while the static algorithm is performing extremely well for this scene from a visual inspection. The discrepancy between accuracy values and the visual results is most likely due to flaws in the ground-truth data. Again, areas that appear as forests in the RGB satellite image are not always correctly classified in forest regions in the ground truth image. It appears that the terrain classification in the ground-truth image is more generalized, and less precise than what can be seen in the satellite images. This, combined with the morphological resizing operation to create images of the same size, and the manual steps necessary to generate a comparable ground-truth image, leaves room for error in the accuracy calculation. Without precise ground-truth data, it is difficult to make assumptions about the quantitative comparison, but the static algorithm may be performing even better than its accuracy value suggests, and there is evidence to suggest it performs better than DetecTree.

\section{Conclusion and Next Steps}
This work sought to identify a static algorithm that could provide a pixel-level classification of vegetation as either forest or non-forest from Sentinel-2 satellite images. Classifiers currently exist for identifying tree pixels in images, but they are designed for aerial photos that are closer to the ground than the high-level Sentinel-2 satellite images being used in this paper. \cite{bosch_detectree_2020} The process identified in this paper shows strong initial results, even with just a basic texture filter and NDVI mask, but more work is necessary to properly quantify its accuracy level. 

As seen in the results section, the ground-truth data used to compare classification processes may lack some inherent accuracy, which diminishes the metrics for the classification processes compared in this paper. Identifying a source of ground-truth data, with label precision at the same level of detail as the satellite images being used, is a critical next step to verify the results of any detection algorithm. Various government agencies maintain detailed terrain inventories, though these can present inaccuracies. \cite{ontario_forest_nodate2} A complete assessment of multiple forest classification processes will require creating a database with various terrain types, including multiple forest types (e.g. temperate, tropical, boreal), labeled with the true terrain classifications. \cite{noauthor_forest_nodate1}

Future work will also require creating a more rigorous feature bank designed specifically for parsing forest areas at the desired field of view and resolution. By supplementing texture features with additional spectral and contextual features, there is potential to further increase accuracy. One example is testing which color bands and visual ratios best leverage the spectral information the satellites capture, which may also include using multiple ratios in a spectral feature bank. To this end, steps remain to fine-tune a highly accurate static algorithm to identify forest areas separately from non-forest vegetation, but initial results show that even a basic combination of textural and spectral features can provide a high-quality classification.

\bibliographystyle{IEEEtran}
\bibliography{IEEEabrv,IEEEPacrim,sentinel_texture}

\end{document}